\documentclass[conference, a4paper]{IEEEtran}

\usepackage[a4paper,
            left=1.6cm,
            right=1.6cm,
            top=2.0cm,
            bottom=4.4cm,
            columnsep=0.7cm  
]{geometry}

\usepackage{setspace}
\IEEEoverridecommandlockouts
\usepackage{amsmath,amssymb,amsfonts}
\usepackage{amsthm}
\usepackage{algorithm}
\usepackage{algorithmic}
\usepackage{array}
\usepackage{booktabs}%
\usepackage{cite}
\usepackage{color}
\usepackage{enumitem}
\usepackage{graphicx}
\usepackage{mathrsfs}
\usepackage{multirow}
\usepackage{stfloats}
\usepackage{subfigure}
\usepackage{tikz}
\usepackage{textcomp}
\usepackage{tabularx}
\usepackage{url}
\usepackage{verbatim}
\usetikzlibrary{arrows.meta, positioning, fit, calc}
\begin{document}
\title{Prototype-Enhanced Multi-View Learning for Thyroid Nodule Ultrasound Classification}

\author{Yangmei Chen,
        Zhongyuan Zhang,
        Xikun Zhang,
        Xinyu Hao, 
        Mingliang Hou\textsuperscript{*},
        Renqiang Luo\textsuperscript{*},
        Ziqi Xu
\thanks{Yangmei Chen is with the College of Software, Jilin University, Changchun 130012, China (chenym5523@mails.jlu.edu.cn).}
\thanks{Zhongyuan Zhang and Renqiang Luo are with the College of Computer Science and Technology, Jilin University, Changchun 130012, China (zhongyuanz25@mails.jlu.edu.cn, lrenqiang@jlu.edu.cn).}
\thanks{Xikun Zhang and Ziqi Xu are with the School of Computing Technologies, RMIT University, Melbourne, VIC 3000, Australia (\{xikun.zhang, ziqi.xu\}@rmit.edu.au).}
\thanks{Xinyu Hao is with the School of Software Technology, Dalian University of Technology, Dalian 116024, China (xihao@dlut.edu.cn).}
\thanks{Mingliang Hou is with the Guangdong Institute of Smart Education, Jinan University, Guangzhou 510632, China (teemohold@outlook.com)}
\thanks{Corresponding author: Mingliang Hou, Renqiang Luo.}}



\maketitle

\begin{abstract}
Thyroid nodule classification using ultrasound imaging is essential for early diagnosis and clinical decision-making; however, despite promising performance on in-distribution data, existing deep learning methods often exhibit limited robustness and generalisation when deployed across different ultrasound devices or clinical environments. 
This limitation is mainly attributed to the pronounced heterogeneity of thyroid ultrasound images, which can lead models to capture spurious correlations rather than reliable diagnostic cues. 
To address this challenge, we propose PEMV-thyroid, a Prototype-Enhanced Multi-View learning framework that accounts for data heterogeneity by learning complementary representations from multiple feature perspectives and refining decision boundaries through a prototype-based correction mechanism with mixed prototype information. 
By integrating multi-view representations with prototype-level guidance, the proposed approach enables more stable representation learning under heterogeneous imaging conditions. 
Extensive experiments on multiple thyroid ultrasound datasets demonstrate that PEMV-thyroid consistently outperforms state-of-the-art methods, particularly in cross-device and cross-domain evaluation scenarios, leading to improved diagnostic accuracy and generalisation performance in real-world clinical settings. 
The source code is available at~\url{https://github.com/chenyangmeii/Prototype-Enhanced-Multi-View-Learning}.

\end{abstract}

\begin{IEEEkeywords}
Thyroid nodule classification, Ultrasound imaging, Multi-view learning
\end{IEEEkeywords}

\section{Introduction}
\par Thyroid nodules are among the most common diseases of the endocrine system and exhibit a high prevalence in the general population~\cite{dean2008epidemiology}. 
Accurate differentiation between benign and malignant nodules is therefore critical for guiding clinical decision-making, reducing unnecessary biopsies, and avoiding excessive invasive treatments~\cite{haugen20162015}. 
Ultrasound imaging is widely adopted as the primary screening modality due to its non-invasive nature, low cost, and real-time capability~\cite{zhou2025segment}. 
However, the visual assessment of thyroid ultrasound images remains highly dependent on clinicians’ subjective interpretation, which can vary across experience levels and clinical settings, leading to inconsistent diagnoses and suboptimal decision-making.

\par In recent years, deep learning techniques have been extensively applied to thyroid nodule diagnosis using ultrasound imaging~\cite{wildman2019using,buda2019management}. 
A wide range of approaches is explored, including dynamic ultrasound video analysis~\cite{qian2025deep}, multimodal deep learning frameworks~\cite{wen2025multimodal}, and hybrid models that integrate traditional machine learning with deep neural networks~\cite{fan2024stable}. 
These methods demonstrate promising performance in improving classification accuracy and diagnostic efficiency~\cite{ziadi2024ai}. 
Moreover, they provide effective technical support for alleviating clinicians’ workload, reducing unnecessary invasive procedures, and enhancing diagnostic consistency in clinical practice~\cite{grani2024thyroid}.

\par
\begin{figure}[t]
\centering
\begin{minipage}[t]{0.32\linewidth}
    \centering
    \includegraphics[width=\linewidth]{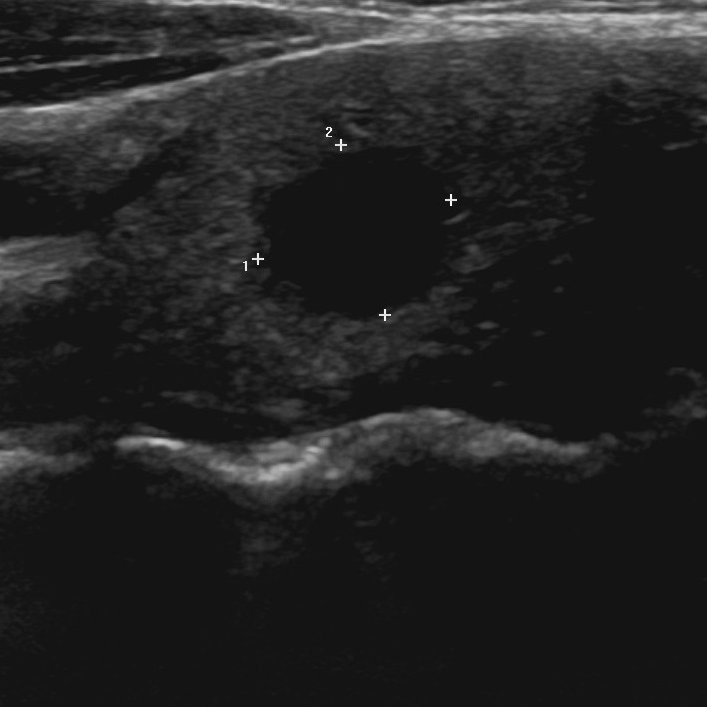}\\[-0.2em]
    \vspace{0.2em}
    {\small 1}
    \end{minipage}
    \hfill
    \begin{minipage}[t]{0.32\linewidth}
    \centering
    \includegraphics[width=\linewidth]{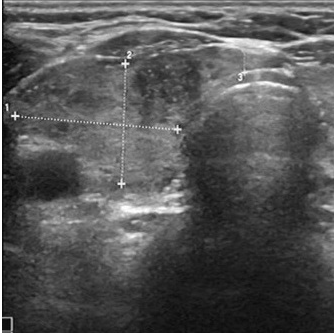}\\[-0.2em]
    \vspace{0.2em}
    {\small 2}
    \end{minipage}
    \hfill
    \begin{minipage}[t]{0.32\linewidth}
    \centering
    \includegraphics[width=\linewidth]{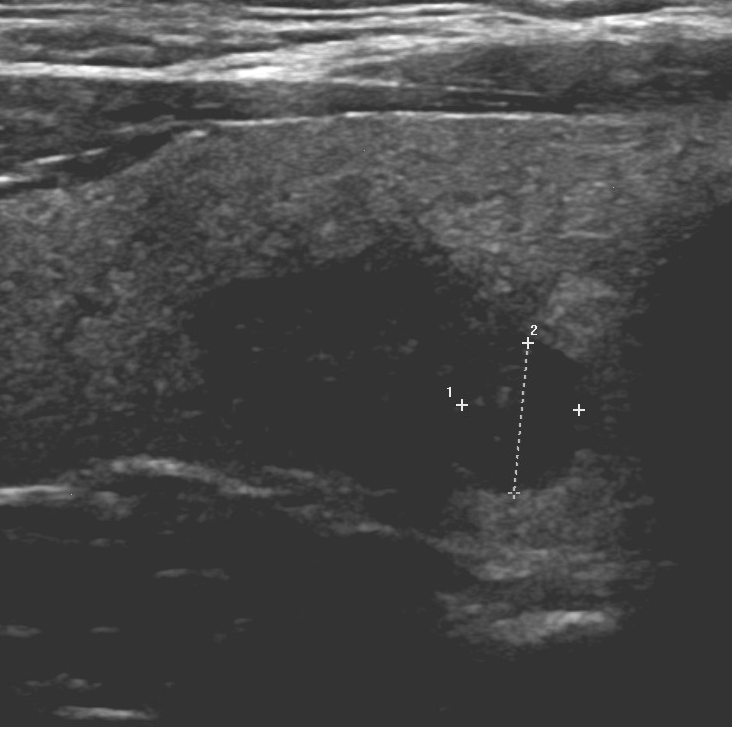}\\[-0.2em]
    \vspace{0.2em}
    {\small 3}
\end{minipage}

\vspace{0.6em}

\resizebox{\linewidth}{!}{%
\begin{tabular}{ccccccc}
    \toprule
    Image & Shape & Direction & Margin & Edge & Internal Echo & Back Echo \\
    \midrule
    1 & Round & Horizontal & Clear & Complete & Hypoechoic & Mixed \\
    2 & Irregular & Horizontal & Unclear & Incomplete & Hyperechoic & Mixed \\
    3 & Oval & Vertical & Unclear & Incomplete & Hypoechoic & Boosted \\
    \bottomrule
\end{tabular}%
}
\caption{Examples illustrating pronounced intra-class heterogeneity in thyroid ultrasound images, where nodules of the same pathological type exhibit diverse visual manifestations across multiple lesion attributes.}
\label{fig:fig1}
\end{figure}

 Despite recent advances, several critical challenges remain unresolved. 
When trained models are deployed on datasets collected from different ultrasound devices or clinical environments, their performance often degrades significantly~\cite{guan2021domain}, indicating limited robustness and poor generalisation. 
Although variance pooling strategies and data augmentation techniques are introduced to mitigate this issue, these approaches remain sensitive to variations in imaging conditions and nodule characteristics, resulting in only marginal performance improvements. 
This limitation is largely attributed to the pronounced heterogeneity of thyroid ultrasound images~\cite{faes2019automated}, which arises from variations in imaging equipment, acquisition protocols, operator expertise, and intrinsic differences in nodule appearance. 
As illustrated in Figure~\ref{fig:fig1}, thyroid nodules sharing the same pathological type can exhibit markedly different visual manifestations in ultrasound images, including variations in echogenicity, margin definition, shape, and internal texture. 
Such pronounced intra-class heterogeneity may induce spurious correlations during model training, causing deep learning models to rely on non-causal visual cues and consequently undermining their robustness and generalisation across diverse clinical settings.

\par To address these challenges, we propose PEMV-thyroid, a Prototype-Enhanced Multi-View Learning framework for thyroid nodule ultrasound classification. 
The proposed approach aims to improve robustness by explicitly accounting for data heterogeneity in the relationship between image representations and diagnostic outcomes. 
It comprises two key components: a Multi-View Feature Extraction (MVFE) module and a Prototype-Based Correction (PBC) module. The MVFE module constructs complementary representations from multiple feature perspectives, while the PBC module refines decision boundaries by incorporating mixed prototype information to reduce the influence of spurious correlations. 
Extensive experiments demonstrate that PEMV-thyroid consistently improves diagnostic accuracy and generalisation performance, underscoring its practical effectiveness for thyroid nodule ultrasound classification. 
In summary, our main contributions are as follows:

\begin{itemize} [leftmargin=0.5cm]
    \item We propose PEMV-thyroid, a Prototype-Enhanced Multi-View learning framework for thyroid nodule ultrasound classification that accounts for data heterogeneity between image representations and diagnostic outcomes, reducing spurious correlations across diverse clinical settings.
    \item We design a prototype-based correction mechanism that integrates multi-view representations with mixed prototype information to enable more stable and reliable learning under heterogeneous imaging conditions.
    \item We conduct extensive experiments on thyroid ultrasound datasets, showing that PEMV-thyroid consistently outperforms state-of-the-art methods, particularly in cross-device and cross-domain scenarios, leading to improved diagnostic accuracy and generalisation.
\end{itemize}

\section{Related Work}
\par Medical image classification aims to automatically predict clinically relevant labels from medical images, thereby providing decision support for disease screening and diagnosis. 
In this work, we focus on benign--malignant classification of thyroid nodules in ultrasound images. 
However, thyroid ultrasound images often exhibit speckle noise, low contrast, and substantial appearance variations across imaging devices and operators, which can hinder model generalisation.

\par Medical image classification has evolved from hand-crafted feature-based methods to deep CNN-based end-to-end learning, and more recently to transformer-based architectures and large-scale pretraining or self-supervised learning paradigms. 
To address common challenges such as domain shift and imaging style variations, existing studies have sought to improve robustness from both data- and representation-level perspectives. 
For example, Mixup~\cite{zhou2021domain} mitigates overfitting by interpolating and mixing training samples, MixStyle~\cite{zhang2018mixup} enhances cross-domain generalisation by perturbing feature statistics, and Fishr~\cite{rame2022fishr} promotes invariant learning through gradient regularisation. 
Nevertheless, these methods may be insufficient for addressing disease heterogeneity and its associated confounding factors, often resulting in suboptimal performance in real-world clinical settings.

\section{Methodology}
\par We address thyroid nodule classification in ultrasound, formulated as a binary prediction problem. 
Let $\mathcal{D}=\{(x_i,y_i)\}_{i=1}^{N}$, where $x_i$ denotes an ultrasound image and $y_i\in\{0,1\}$ is its label ($0$: benign, $1$: malignant). 
While conventional classifiers optimise the observational objective associated with $p_\theta(y\mid x)$, the proposed PEMV-thyroid framework is motivated by the presence of unmeasured confounding and aims to learn more stable predictive relationships guided by causal principles.
Specifically, PEMV-thyroid constructs multi-view feature representations as an intermediate mediator $A$ through a Multi-View Feature Extraction (MVFE) module, and subsequently refines this mediator via a prototype-based correction mechanism to obtain $\hat{A}$, which is inspired by the front-door adjustment concept~\cite{pearl2000causality,XuCLL0Y24} to attenuate confounder-induced variations without explicitly modelling unobserved confounders. 
The final prediction is produced by feeding the concatenation of a global feature $g$ and the refined mediator $\hat{A}$ into a classifier head:
\begin{align}
p_{\theta}(y \mid g,\hat{A}) &= \mathrm{softmax}\big(f_c([g;\hat{A}])\big), \\
\hat{y} &= \arg\max_{c\in\{0,1\}} p_{\theta}(y=c \mid g,\hat{A}).
\end{align}

\par
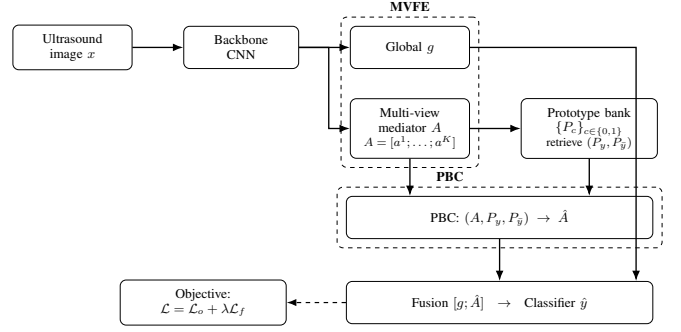
\begin{figure}[t]
    \centering
    \resizebox{\linewidth}{!}{%
    \begin{tikzpicture}[
      font=\small,
      >=Latex,
      node distance=7mm and 12mm,
      box/.style={draw, rounded corners, align=center, inner sep=6pt, minimum height=10mm},
      gbox/.style={draw, rounded corners, dashed, inner sep=6pt},
      arrow/.style={->, line width=0.9pt},
      darrow/.style={->, dashed, line width=0.8pt}
    ]
    \usetikzlibrary{calc,fit}
    
    \node[box, minimum width=2.8cm] (x) {Ultrasound\\image $x$};
    \node[box, right=of x, minimum width=2.7cm] (bb) {Backbone\\CNN};
    \draw[arrow] (x) -- (bb);
    
    \node[box, right=12mm of bb, minimum width=2.8cm] (g) {Global $g$};
    \node[box, below=7mm of g, minimum width=2.8cm] (A)
    {Multi-view\\mediator $A$\\{\footnotesize $A=[a^1;\dots;a^K]$}};
    \node[gbox, fit=(g)(A), label={[font=\small]above:\textbf{MVFE}}] (mvfe) {};
    
    \draw[arrow] (bb.east) -- ++(7mm,0) |- (g.west);
    \draw[arrow] (bb.east) -- ++(7mm,0) |- (A.west);
    
    \node[box, right=12mm of A, minimum width=3.2cm] (pb)
    {Prototype bank\\$\{P_c\}_{c\in\{0,1\}}$\\{\footnotesize retrieve $(P_y,P_{\bar y})$}};
    \draw[arrow] (A.east) -- (pb.west);
    
    \coordinate (midAP) at ($(A)!0.5!(pb)$);
    \node[box, below=16mm of midAP, minimum width=7.2cm] (pbc_core)
    {PBC: $(A,P_y,P_{\bar y}) \;\rightarrow\; \hat{A}$};
    \node[gbox, fit=(pbc_core), label={[font=\small]above left:\textbf{PBC}}] (pbc) {};
    
    \coordinate (inA) at ($(A.south |- pbc_core.north)$);
    \coordinate (inB) at ($(pb.south |- pbc_core.north)$);
    \draw[arrow] (A.south) -- (inA);
    \draw[arrow] (pb.south) -- (inB);
    
    \node[box, below=10mm of pbc_core, minimum width=7.2cm] (clf)
    {Fusion $[g;\hat{A}]$ \; $\rightarrow$ \; Classifier $\hat{y}$};
    
    \draw[arrow] (pbc_core.south) -- (clf.north);
    
    \coordinate (g_in) at ($(clf.north east) + (-4mm,0)$);
    \draw[arrow] (g.east) -| (g_in);
    
    \node[box, left=14mm of clf, minimum width=3.9cm] (loss)
    {Objective:\\ $\mathcal{L}=\mathcal{L}_o+\lambda\mathcal{L}_f$};
    \draw[darrow] (clf.west) -- (loss.east);
    
    \end{tikzpicture}%
    }
    \caption{Overview of the proposed PEMV-thyroid framework for thyroid ultrasound classification. The MVFE module extracts multi-view mediator representations, while the PBC module refines these representations using class-conditional prototypes to mitigate spurious correlations under heterogeneous imaging conditions.}
    \label{fig:mvfe_pbc_overview}
\end{figure}

 During training, we optimise a joint objective that combines the standard classification loss with an additional fusion loss to jointly supervise representation learning and prototype-based correction. 
The overall architecture of the proposed method is shown in Figure~\ref{fig:mvfe_pbc_overview}.

\subsection{Front-door Adjustment}
\par A major difficulty in thyroid ultrasound classification is that acquisition-related factors (e.g., device settings and operator-dependent scanning) may introduce latent confounding that affects both the observed image appearance $x$ and the diagnostic label $y$. 
As a result, directly fitting the observational conditional $p(y\mid x)$ can be unstable across domains. 
From a causal perspective, introducing an intermediate representation that captures disease-relevant evidence transmitted from the image to the label can help attenuate confounder-induced spurious correlations. 
In this work, PEMV-thyroid adopts such an intermediate representation $A$ as a mediator, inspired by the front-door adjustment principle. 

\par Under the front-door assumptions, the interventional effect can be expressed using only observational quantities as:
\begin{align}
    p(y \mid do(x))
    &= \sum_{a} p(a\mid x)\, \sum_{x'} p(y \mid a, x')\, p(x').
    \label{eq:frontdoor}
\end{align}

\par Eq.~\eqref{eq:frontdoor} suggests a decomposition into two stages: 
(i) learning how the image gives rise to an intermediate representation, i.e., $p(a\mid x)$, and 
(ii) estimating the label distribution conditioned on this representation while marginalising over the image distribution. 
In the following, we describe how PEMV-thyroid instantiates the mediator $A$ using multi-view feature representations and how a prototype-based correction mechanism is employed to approximate the intervention-inspired effect implied by Eq.~\eqref{eq:frontdoor}.

\subsection{Instantiating the mediator via multi-view representations}
\par We implement the mediator-generation term $p(a\mid x)$ by extracting disease-related representations from the input ultrasound image. 
Specifically, given an image $x$, a backbone network produces a shared feature map, from which we derive 
(i) a global representation $g$ that summarises holistic semantics, and 
(ii) a set of $K$ view-specific representations $\{a_k\}_{k=1}^{K}$ that capture complementary evidence. 
These view-specific features are treated as the mediator, and the aggregated mediator is defined as
\begin{equation}
A = \big[a_1;\, a_2;\, \ldots;\, a_K\big],
\end{equation}
where $[\cdot;\cdot]$ denotes concatenation.

\par The multi-view design is particularly well suited to thyroid ultrasound imaging, where speckle noise, low contrast, and device- or operator-dependent appearance variations can induce spurious shortcuts when relying solely on global features. 
By decomposing disease evidence into multiple complementary views, the mediator $A$ encourages the model to encode more structured and reusable representations, which subsequently facilitates robustness-oriented correction under heterogeneous imaging conditions.

\subsection{Prototype-based correction of the mediator}
\par To mitigate the influence of unmeasured confounding on the learned mediator, PEMV-thyroid incorporates a prototype-based correction mechanism that refines mediator representations using class-conditional reference patterns. 
For each class $c\in\{0,1\}$, we maintain a mediator prototype $P_c$, which serves as a class-specific reference representation. 
In practice, each prototype is updated during training by aggregating mediator features from samples belonging to class $c$, yielding a stable estimate of typical disease-related patterns for that class.

\par Given a training sample $(x,y)$, we retrieve the corresponding same-class prototype $P_y$ and additionally sample a different-class prototype $P_{\bar{y}}$. 
These prototypes are jointly leveraged to refine the mediator extracted from $x$, producing a corrected mediator $\hat{A}$. 
Intuitively, the same-class prototype encourages alignment with class-relevant evidence, while the different-class prototype provides complementary contrast that discourages reliance on confounder-driven shortcuts. 
Through this refinement process, the corrected mediator becomes more invariant to acquisition-related variations, thereby improving robustness across devices and clinical environments.

\par After obtaining the corrected mediator $\hat{A}$, it is fused with the global representation $g$ for final classification. 
The model is trained using a joint objective that combines the standard classification loss with an additional fusion loss, which jointly supervises representation learning and prototype-based correction under heterogeneous imaging conditions.

\subsection{Fusion and learning objective}
\par With the corrected mediator $\hat{A}$, PEMV-thyroid performs prediction by jointly leveraging global and mediator-level evidence. 
Specifically, we concatenate the global representation $g$ with the corrected mediator to form a fused feature
$z = [g;\hat{A}]$, which is fed into a classifier head $f_c$ to produce logits and the predictive distribution
$p_{\theta}(y\mid z)=\mathrm{softmax}(f_c(z))$.

\par The model is trained using a joint learning objective. 
The first term, $\mathcal{L}_o$, is the standard cross-entropy loss that enforces discriminative learning on the training set. 
However, optimising $\mathcal{L}_o$ alone may encourage the model to exploit spurious correlations that are predictive only under specific acquisition conditions. 
To further promote robustness under heterogeneous imaging environments, PEMV-thyroid introduces an additional fusion loss $\mathcal{L}_f$, which provides complementary supervision for the corrected mediator and its fusion with the global representation, encouraging more stable and invariant decision cues.

\par The overall optimisation objective is given by
\begin{equation}
\mathcal{L} = \mathcal{L}_o + \lambda\,\mathcal{L}_f,
\end{equation}

\begin{equation}
\mathcal{L}_{o}
= -\frac{1}{N}\sum_{i=1}^{N}\sum_{c=1}^{C} y_{ic}\,
\log\left(\frac{\exp(\hat{y}_{ic})}{\sum_{j=1}^{C}\exp(\hat{y}_{ij})}\right),
\label{eq:Lo}
\tag{6}
\end{equation}

\begin{equation}
\mathcal{L}_{f}
= -\sum_{x'} P(x')\Bigg[
\begin{aligned}[t]
& P(\hat{y}^{c})\, l_{c}\,\log\frac{\exp(\hat{y}^{c})}{\sum_{j=1}^{C}\exp(\hat{y}^{j})} \\
& + P(\hat{y}^{c'})\, l_{c'}\,\log\frac{\exp(\hat{y}^{c'})}{\sum_{j=1}^{C}\exp(\hat{y}^{j})}
\end{aligned}
\Bigg],
\label{eq:Lf}
\tag{7}
\end{equation}
where $\lambda$ controls the relative contribution of the fusion loss.

\section{Experiments}
\subsection{Datasets}
\par In this study, we evaluate the proposed method on two publicly available thyroid ultrasound image datasets, namely TN$5000$ and TN$3$K. 
Both datasets are designed for thyroid nodule analysis and support a binary classification task of distinguishing benign and malignant nodules. 
They are selected for their clinical relevance, annotated diagnostic labels, and diversity of imaging conditions, which together enable a comprehensive evaluation of model robustness and generalisation.

\begin{itemize} [leftmargin=0.5cm]
    \item TN$5000$: A thyroid ultrasound image dataset in which each image is annotated with a benign or malignant diagnostic label. 
    The dataset contains images acquired under diverse clinical conditions, including variations in ultrasound devices, imaging parameters, and nodule appearances, providing a realistic benchmark for evaluating robustness and generalisation performance.
    \item TN$3$K: A publicly available thyroid ultrasound dataset annotated with benign and malignant labels. 
    As ultrasound is a primary non-invasive modality for thyroid nodule assessment, TN$3$K has strong clinical relevance for computer-aided diagnosis research and poses additional challenges due to variations in acquisition settings and device configurations.
\end{itemize}

\par Following standard practices in medical image classification, all ultrasound images are resized to a fixed resolution and normalised before being fed into the network. 
Images in both datasets are divided into disjoint training, validation, and test sets, which are used for model optimisation, hyperparameter selection, and final performance evaluation, respectively.

\par TN$5000$ consists of $5{,}000$ images with predefined splits following the PASCAL VOC protocol, including $3{,}500$ training, $500$ validation, and $1{,}000$ test images (approximately $70$\%/$10$\%/$20$\%). 
We strictly follow these official splits and convert the original detection annotations into image-level binary labels without altering the data partitioning. 
TN$3$K contains $3{,}493$ images with an official test set of $614$ images, while the remaining $2{,}879$ images are split into training and validation sets using an $8{:}2$ ratio, resulting in $2{,}303$ training and $576$ validation images (approximately $66$\%/$16$\%/$18$\%).

\par During training, data augmentation is applied only to the training images to improve model generalisation, while no augmentation is used for validation or test samples. 
All data splits are fixed and specified via predefined text files to ensure reproducibility across experiments.

\subsection{Baselines}
\par To validate the effectiveness of PEMV-thyroid for thyroid ultrasound image classification, we compare it with several representative and reproducible baseline methods that are widely adopted in medical image analysis. 
All methods are trained and evaluated under the same data splits, input preprocessing procedures, and evaluation metrics to ensure a fair comparison.

\par We consider the following baseline methods:
\begin{itemize} [leftmargin=0.5cm]
    \item ResNet$18$ (ERM)~\cite{he2016deep}: A standard convolutional neural network trained with empirical risk minimisation is adopted as the primary backbone baseline. 
    This setting serves as a strong and widely used reference for binary thyroid nodule classification.
    \item Fishr~\cite{rame2022fishr}: Fishr is an invariant feature learning method that regularises the variance of gradients across environments to reduce reliance on spurious correlations. 
    In our implementation, Fishr is applied as an additional regularisation term on top of the backbone training objective to enhance robustness under heterogeneous imaging conditions.
    \item MixStyleNet~\cite{zhang2018mixup}: MixStyleNet performs feature-level style perturbation by mixing channel-wise statistics, such as mean and variance, during training. 
    This strategy simulates domain and style shifts caused by different ultrasound devices and acquisition settings, making it particularly relevant for ultrasound images with substantial appearance variability.
    \item MixupNet~\cite{zhou2021domain}: MixupNet applies the Mixup strategy to construct virtual training samples by linearly interpolating pairs of input images and their corresponding labels.
    This regularisation encourages smoother decision boundaries and is commonly used to improve generalisation in medical image classification.
\end{itemize}

\par These baselines represent commonly adopted strategies for improving robustness and generalisation in medical image classification, including empirical risk minimisation, data augmentation, and invariant representation learning. 
By evaluating PEMV-thyroid against Fishr, MixStyleNet, and MixupNet under a unified experimental protocol, we provide a systematic comparison with methods that address domain variability and spurious correlations from different perspectives.

\subsection{Experimental Setup}
\par All experiments are conducted on a workstation equipped with an NVIDIA L$40$ GPU. 
The software environment includes Python $3.8.20$, PyTorch $1.10.1$, and CUDA $11.3$. 
ResNet$18$ is adopted as the backbone network for all methods. 
All thyroid ultrasound images are resized to $128 \times 128$ pixels. 
For both TN$5000$ and TN$3$K datasets, models are trained using the AdamW optimizer with an initial learning rate of $1 \times 10^{-4}$ and a batch size of $16$. 
All reported results are obtained by averaging over five runs with different random seeds.

\subsection{Main Results}
\par In this section, we present a comprehensive evaluation of PEMV-thyroid against state-of-the-art baselines across two real-world thyroid nodule ultrasound datasets, TN$3$K and TN$5000$. 
The comparison focuses on four commonly used metrics, including accuracy (ACC), precision (P), recall (R), and F$1$-score (F$1$). 
Overall, the quantitative results reported in Table~\ref{tab:tn3k_results} and Table~\ref{tab:tn5000_results} show that PEMV-thyroid consistently outperforms existing methods, demonstrating its effectiveness in learning robust representations for thyroid nodule classification.

\par On the TN$3$K dataset, PEMV-thyroid achieves clear improvements over all baseline methods, as summarised in Table~\ref{tab:tn3k_results}. 
Specifically, compared with MixupNet, which constructs virtual training samples via linear interpolation, PEMV-thyroid yields improvements of $3.97$\%, $2.64$\%, $10.51$\%, and $7.38$\% in accuracy, precision, recall, and F$1$-score, respectively. 
Notably, PEMV-thyroid achieves a substantial gain in recall ($60.76$\% $\rightarrow$ $71.27$\%), which is particularly important in clinical diagnosis where missing malignant cases should be minimised. 
Moreover, PEMV-thyroid attains an ACC of $82.08$\% and an F$1$-score of $75.32$\%, outperforming the strongest baseline Fishr (ACC $79.74$\%, F$1$ $71.71$\%). 
These results indicate that PEMV-thyroid better mitigates the impact of data heterogeneity and reduces reliance on spurious correlations, leading to improved generalisation under challenging imaging conditions.

\par On the TN$5000$ dataset, all methods achieve relatively high performance, suggesting a more stable training distribution. 
As shown in Table~\ref{tab:tn5000_results}, PEMV-thyroid delivers the best overall performance, achieving $86.50$\% ACC and $90.99$\% F$1$-score, compared with the strongest baseline Fishr ($85.82$\% ACC, $90.55$\% F$1$). 
These results demonstrate that PEMV-thyroid not only excels on more heterogeneous data such as TN$3$K, but also delivers consistent performance gains on TN$5000$, highlighting its robustness across different thyroid ultrasound datasets.

\begin{table}[t]
    \centering
    \caption{Comparison of different methods on the TN$3$K dataset. All results are reported in percentage (\%), and the best performance is highlighted in bold.}
    \label{tab:tn3k_results}
    \small
    \setlength{\tabcolsep}{2.5pt}
    \renewcommand{\arraystretch}{1}
    \begin{tabular}{lcccc}
    \toprule
    Method & ACC(\%) & P(\%) & R(\%) & F1(\%) \\
    \midrule
    ResNet$18$
    & $79.67_{\pm 1.96}$ & $80.88_{\pm 5.00}$ & $62.29_{\pm 4.22}$ & $70.17_{\pm 2.76}$ \\
    Fishr 
    & $79.74_{\pm 3.02}$ & $77.61_{\pm 5.87}$ & $67.88_{\pm 9.78}$ & $71.71_{\pm 5.45}$ \\
    MixupNet 
    & $78.11_{\pm 2.86}$ & $77.31_{\pm 3.42}$ & $60.76_{\pm 6.41}$ & $67.94_{\pm 5.16}$ \\
    MixStyleNet 
    & $78.96_{\pm 1.54}$ & $76.12_{\pm 2.13}$ & $66.02_{\pm 4.21}$ & $70.63_{\pm 2.71}$ \\
    \midrule
    {PEMV-thyroid} 
    & $\textbf{82.08}_{\pm 1.14}$ & $\textbf{79.95}_{\pm 1.11}$ & $\textbf{71.27}_{\pm 3.23}$ & $\textbf{75.32}_{\pm 2.04}$ \\
    \bottomrule
    \end{tabular}%
\end{table}

\begin{table}[t]
    \centering
    \caption{Comparison of different methods on the TN$5000$ dataset. All results are reported in percentage (\%), and the best performance is highlighted in bold.}
    \label{tab:tn5000_results}
    \small
    \setlength{\tabcolsep}{2.5pt}
    \renewcommand{\arraystretch}{1}
    \begin{tabular}{lcccc}
    \toprule
    Method & ACC(\%) & P(\%) & R(\%) & F1(\%) \\
    \midrule
    ResNet18
    & $85.68_{\pm 0.66}$ & $88.78_{\pm 1.42}$ & $92.09_{\pm 1.37}$ & $90.39_{\pm 0.39}$ \\
    Fishr 
    & $85.82_{\pm 0.46}$ & $88.28_{\pm 1.04}$ & $92.97_{\pm 1.01}$ & $90.55_{\pm 2.06}$ \\
    MixupNet 
    & $85.66_{\pm 0.60}$ & $88.75_{\pm 0.48}$ & $92.07_{\pm 1.52}$ & $90.37_{\pm 0.50}$ \\
    MixStyleNet 
    & $84.68_{\pm 0.79}$ & $87.52_{\pm 1.19}$ & $92.23_{\pm 1.41}$ & $89.80_{\pm 0.52}$ \\
    \midrule
    {PEMV-thyroid} 
    & $\textbf{86.50}_{\pm 0.55}$ & $\textbf{88.88}_{\pm 0.87}$ & $\textbf{93.21}_{\pm 0.97}$ & $\textbf{90.99}_{\pm 0.36}$ \\
    \bottomrule
    \end{tabular}%
\end{table}

\subsection{Sensitivity Analysis}
\par
\begin{figure}[t]
    \centering
    \includegraphics[width=1.0\linewidth]{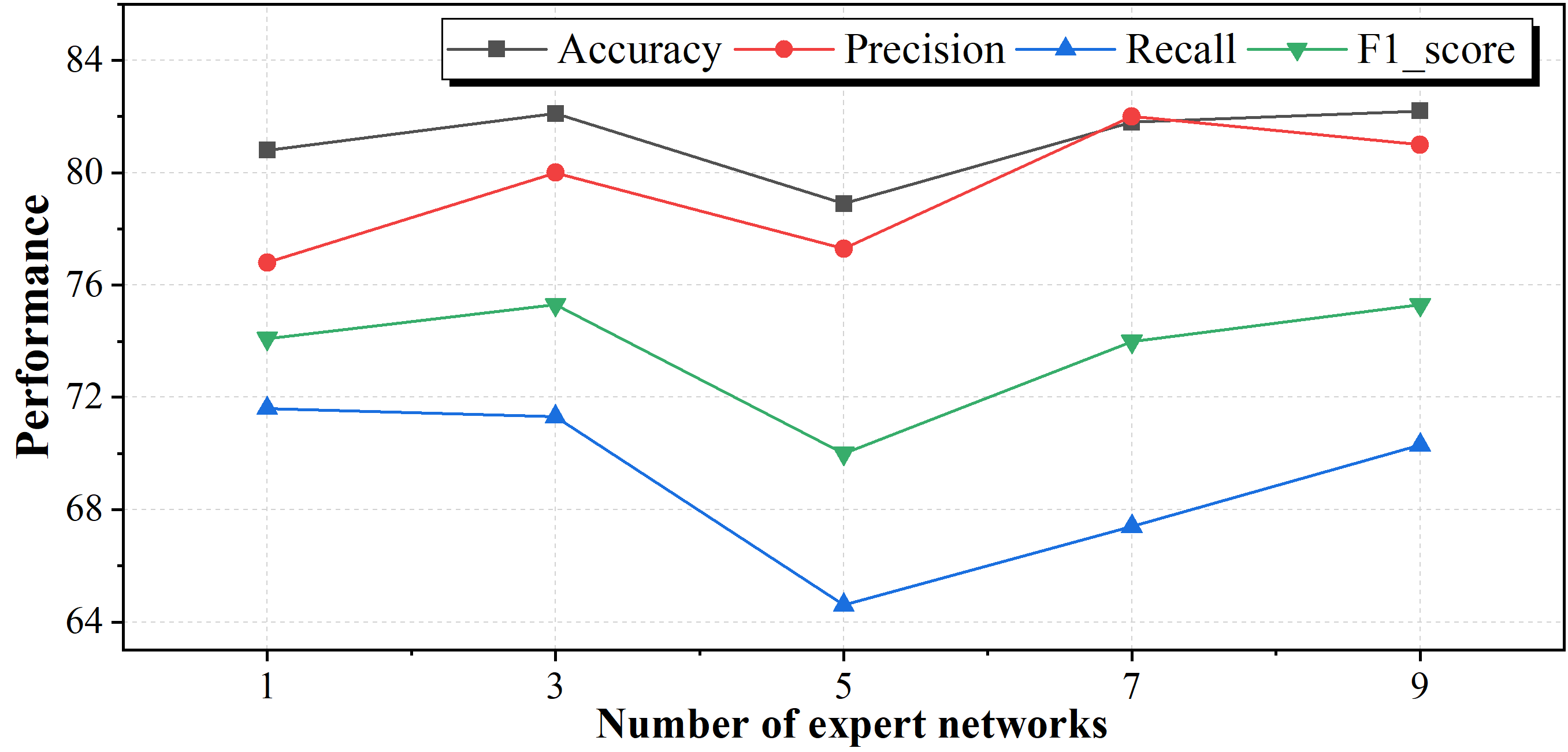}
    \caption{Sensitivity analysis of the number of expert networks (\textit{num\_att}) in the MVFE module on the TN$3$K dataset, evaluated using ACC, P, R, and F$1$ (\%).}
    \label{fig:num_att_tn3k}
\end{figure}

 We analyse the effect of the number of expert networks in the MVFE module by varying \textit{num\_att} from $1$ to $9$ on the TN$3$K dataset (Fig.~\ref{fig:num_att_tn3k}). 
Overall, the performance is sensitive to the choice of \textit{num\_att} but remains relatively stable within a reasonable range. 
Among all configurations, \textit{num\_att}$=3$ achieves the best overall performance, with $82.08$\% ACC, $79.95$\% precision, $71.27$\% recall, and $75.32$\% F$1$-score. 
Increasing the number of experts beyond this setting does not lead to consistent improvements; for example, \textit{num\_att}$=5$ results in a noticeable performance drop ($78.9$\% ACC and $64.6$\% recall), suggesting that an excessive number of experts may introduce optimisation difficulty or overfitting under limited training data. 
Based on these observations, we adopt \textit{num\_att}$=3$ as the default configuration in all experiments.

\subsection{Ablation Study}
\par We conduct a step-wise ablation study from AB1 to AB5 on the TN$3$K and TN$5000$ datasets, with mean$\pm$std results reported in Table~\ref{tab:ablation}. 
Introducing the multi-view feature extractor (AB$2$) consistently improves the ERM baseline (AB$1$) on both datasets, yielding gains on TN$3$K in ACC ($79.67$\% $\rightarrow$ $79.87$\%) and F$1$ ($70.17$\% $\rightarrow$ $71.42$\%), and similar improvements on TN$5000$. 
Adding the prototype-based correction module (AB$3$) further boosts recall on TN$3$K from $65.51$\% to $70.25$\%, leading to a higher F$1$-score ($71.69$\%), while the improvement on TN$5000$ remains marginal, reflecting its more stable data distribution. 
Incorporating the information-purity factor alone (AB$4$) causes noticeable performance fluctuations on TN$3$K, particularly a drop in recall to $62.37$\%, indicating that a single constraint is insufficient for stable optimisation. 
By jointly integrating all components, the full model (AB$5$) achieves the best overall performance on both datasets, with TN$3$K reaching $82.08$\% ACC and $75.32$\% F$1$, and TN$5000$ achieving $86.50$\% ACC and $90.99$\% F$1$, demonstrating the complementarity and effectiveness of the proposed framework.

\begin{table}[t]
    \centering
    \caption{Ablation results on the TN$3$K and TN$5000$ datasets (mean$\pm$std, \%). AB$1$: ResNet$18$ (ERM baseline); AB$2$: AB1+MVFE; AB$3$: AB$2$+PBC; AB$4$: AB$3$+IP; AB$5$: full model. The best and second-best results are highlighted in bold and underlined, respectively.}
    \label{tab:ablation}
    \small
    \setlength{\tabcolsep}{3pt}
    \renewcommand{\arraystretch}{1}
    \begin{tabular}{lcccc}
    \toprule
    \multicolumn{5}{c}{{TN$3$K}}\\
    \midrule
    Method & ACC(\%) & P(\%) & R(\%) & F1(\%) \\
    \midrule
    AB$1$ & $79.67_{\pm 1.96}$ & $\textbf{80.88}_{\pm 5.00}$ & $62.29_{\pm 4.22}$ & $70.17_{\pm 2.76}$ \\
    AB$2$ & \underline{$79.87_{\pm 1.81}$} & $79.06_{\pm 4.75}$ & $65.51_{\pm 4.43}$ & $71.42_{\pm 2.36}$ \\
    AB$3$ & $78.73_{\pm 1.57}$ & $73.80_{\pm 4.41}$ & \underline{$70.25_{\pm 5.44}$} & \underline{$71.69_{\pm 2.09}$} \\
    AB$4$ & $78.73_{\pm 1.44}$ & $78.85_{\pm 5.54}$ & $62.37_{\pm 8.88}$ & $68.99_{\pm 3.71}$ \\
    AB$5$ & $\textbf{82.08}_{\pm 1.14}$ & \underline{$79.95_{\pm 1.11}$} & $\textbf{71.27}_{\pm 3.23}$ & $\textbf{75.32}_{\pm 2.04}$ \\
    \midrule
    \multicolumn{5}{c}{{TN5000}}\\
    \midrule
    Method & ACC(\%) & P(\%) & R(\%) & F1(\%) \\
    \midrule
    AB$1$ & $85.68_{\pm 0.66}$ & $88.78_{\pm 1.42}$ & $92.09_{\pm 1.37}$ & $90.39_{\pm 0.39}$ \\
    AB$2$ & $86.10_{\pm 0.77}$ & $88.66_{\pm 0.83}$ & \underline{$92.89_{\pm 1.60}$} & \underline{$90.71_{\pm 0.58}$} \\
    AB$3$ & $85.06_{\pm 0.66}$ & $87.87_{\pm 0.82}$ & $92.34_{\pm 1.80}$ & $90.03_{\pm 0.54}$ \\
    AB$4$ & \underline{$86.12_{\pm 0.73}$} & $\textbf{89.14}_{\pm 0.48}$ & $92.26_{\pm 1.10}$ & $90.67_{\pm 0.54}$ \\
    AB$5$ & $\textbf{86.50}_{\pm 0.55}$ & \underline{$88.88_{\pm 0.87}$} & $\textbf{93.21}_{\pm 0.97}$ & $\textbf{90.99}_{\pm 0.36}$ \\
    \bottomrule
    \end{tabular}%
\end{table}

\section{Conclusion}
\par This work presents PEMV-thyroid, a prototype-enhanced multi-view learning framework for robust thyroid nodule ultrasound classification. 
By explicitly accounting for data heterogeneity through complementary multi-view representations and a prototype-based correction mechanism, the proposed approach mitigates the influence of spurious correlations arising from variations in imaging devices, acquisition protocols, and nodule appearances. 
Extensive experiments on two publicly available thyroid ultrasound datasets demonstrate that PEMV-thyroid consistently outperforms state-of-the-art baselines, with particularly notable improvements under cross-device and heterogeneous settings. 
These results highlight the effectiveness of integrating multi-view representation learning with prototype-guided refinement for improving robustness and generalisation in medical image classification. 
Future work explores extending the proposed framework to other ultrasound-based diagnostic tasks and investigating its applicability to additional medical imaging modalities with pronounced domain variability.

\bibliographystyle{IEEEtran}
\bibliography{NeiGAD}

@article{buda2019management,
  title={Management of thyroid nodules seen on US images: deep learning may match performance of radiologists},
  author={Buda, Mateusz and Wildman-Tobriner, Benjamin and Hoang, Jenny K and Thayer, David and Tessler, Franklin N and Middleton, William D and Mazurowski, Maciej A},
  journal={Radiology},
  volume={292},
  number={3},
  pages={695--701},
  year={2019}
}

@article{dean2008epidemiology,
  title={Epidemiology of thyroid nodules},
  author={Dean, Diana S and Gharib, Hossein},
  journal={Best practice \& research Clinical endocrinology \& metabolism},
  volume={22},
  number={6},
  pages={901--911},
  year={2008}
}

@article{faes2019automated,
  title={Automated deep learning design for medical image classification by health-care professionals with no coding experience: A feasibility study},
  author={Faes, Livia and Wagner, Siegfried K and Fu, Dun Jack and Liu, Xiaoxuan and Korot, Edward and Ledsam, Joseph R and Back, Trevor and Chopra, Reena and Pontikos, Nikolas and Kern, Christoph and others},
  journal={The Lancet Digital Health},
  volume={1},
  number={5},
  pages={e232--e242},
  year={2019}
}

@article{fan2024stable,
  title={Stable Cox regression for survival analysis under distribution shifts},
  author={Fan, Shaohua and Xu, Renzhe and Dong, Qian and He, Yue and Chang, Cheng and Cui, Peng},
  journal={Nature Machine Intelligence},
  volume={6},
  number={12},
  pages={1525--1541},
  year={2024}
}

@article{grani2024thyroid,
  title={Thyroid nodules: diagnosis and management},
  author={Grani, Giorgio and Sponziello, Marialuisa and Filetti, Sebastiano and Durante, Cosimo},
  journal={Nature Reviews Endocrinology},
  volume={20},
  number={12},
  pages={715--728},
  year={2024}
}

@article{guan2021domain,
  title={Domain adaptation for medical image analysis: a survey},
  author={Guan, Hao and Liu, Mingxia},
  journal={IEEE Transactions on Biomedical Engineering},
  volume={69},
  number={3},
  pages={1173--1185},
  year={2021}
}

@article{haugen20162015,
  title={2015 American Thyroid Association management guidelines for adult patients with thyroid nodules and differentiated thyroid cancer: the American Thyroid Association guidelines task force on thyroid nodules and differentiated thyroid cancer},
  author={Haugen, Bryan R and Alexander, Erik K and Bible, Keith C and Doherty, Gerard M and Mandel, Susan J and Nikiforov, Yuri E and Pacini, Furio and Randolph, Gregory W and Sawka, Anna M and Schlumberger, Martin and others},
  journal={Thyroid},
  volume={26},
  number={1},
  pages={1--133},
  year={2016}
}

@inproceedings{he2016deep,
  author       = {Kaiming He and
                  Xiangyu Zhang and
                  Shaoqing Ren and
                  Jian Sun},
  title        = {Deep Residual Learning for Image Recognition},
  booktitle    = {{IEEE} Conference on Computer Vision and Pattern Recognition,
                  {CVPR}},
  pages        = {770--778},
  year         = {2016}
}

@book{pearl2000causality,
  title     = {Causality: Models, Reasoning, and Inference},
  author    = {Pearl, Judea},
  year      = {2000},
  publisher = {Cambridge University Press},
  address   = {Cambridge, UK}
}

@article{qian2025deep,
  title={Deep learning based analysis of dynamic video ultrasonography for predicting cervical lymph node metastasis in papillary thyroid carcinoma},
  author={Qian, Tingting and Zhou, Yahan and Yao, Jincao and Ni, Chen and Asif, Sohaib and Chen, Chen and Lv, Lujiao and Ou, Di and Xu, Dong},
  journal={Endocrine},
  volume={87},
  number={3},
  pages={1060--1069},
  year={2025}
}

@inproceedings{rame2022fishr,
  author       = {Alexandre Ram{\'{e}} and
                  Corentin Dancette and
                  Matthieu Cord},
  title        = {Fishr: Invariant Gradient Variances for Out-of-Distribution Generalization},
  booktitle    = {International Conference on Machine Learning, {ICML}},
  pages        = {18347--18377},
  year         = {2022}
}

@article{wen2025multimodal,
  title={Multimodal model enhances qualitative diagnosis of hypervascular thyroid nodules: Integrating radiomics and deep learning features based on B-mode and PDI images},
  author={Wen, Wen and Zhang, Tingrui and Zhao, Haina and Liu, Jingyan and Jiang, Heng and He, Yushuang and Jiang, Zekun},
  journal={Gland Surgery},
  volume={14},
  number={8},
  pages={1558--1571},
  year={2025}
}

@article{wildman2019using,
  title={Using artificial intelligence to revise ACR TI-RADS risk stratification of thyroid nodules: diagnostic accuracy and utility},
  author={Wildman-Tobriner, Benjamin and Buda, Mateusz and Hoang, Jenny K and Middleton, William D and Thayer, David and Short, Ryan G and Tessler, Franklin N and Mazurowski, Maciej A},
  journal={Radiology},
  volume={292},
  number={1},
  pages={112--119},
  year={2019}
}

@inproceedings{zhang2018mixup,
  author       = {Hongyi Zhang and
                  Moustapha Ciss{\'{e}} and
                  Yann N. Dauphin and
                  David Lopez{-}Paz},
  title        = {mixup: Beyond Empirical Risk Minimization},
  booktitle    = {6th International Conference on Learning Representations, {ICLR}},
  year         = {2018}
}

@inproceedings{zhou2021domain,
  author       = {Kaiyang Zhou and
                  Yongxin Yang and
                  Yu Qiao and
                  Tao Xiang},
  title        = {Domain Generalization with MixStyle},
  booktitle    = {9th International Conference on Learning Representations, {ICLR}},
  year         = {2021}
}

@article{zhou2025segment,
  title={Segment anything model for fetal head-pubic symphysis segmentation in intrapartum ultrasound image analysis},
  author={Zhou, Zihao and Lu, Yaosheng and Bai, Jieyun and Campello, Victor M and Feng, Fan and Lekadir, Karim},
  journal={Expert Systems with Applications},
  volume={263},
  pages={125699},
  year={2025}
}

@inproceedings{ziadi2024ai,
  title={{AI} and {I}o{T} Users, Challenges and Opportunities for e-Health: A Review},
  author={Ziadi, Faten and Fourati, Hend and Saidane, Leila Azouz},
  booktitle={Proceedings of the 2024 International Wireless Communications and Mobile Computing},
  year={2024}
}

@inproceedings{XuCLL0Y24,
  author       = {Ziqi Xu and
                  Debo Cheng and
                  Jiuyong Li and
                  Jixue Liu and
                  Lin Liu and
                  Kui Yu},
  title        = {Causal Inference with Conditional Front-Door Adjustment and Identifiable
                  Variational Autoencoder},
  booktitle    = {The Twelfth International Conference on Learning Representations,
                  {ICLR}},
  year         = {2024}
}

\end{document}